\documentclass[journal]{IEEEtran}

\ifCLASSOPTIONcompsoc
  \usepackage[nocompress]{cite}
\else
  \usepackage{cite}
\fi
\usepackage{booktabs} 
\usepackage{float}
\usepackage{graphics,graphicx}

\usepackage{mathtools}
\usepackage{amsmath}
\usepackage{amsfonts}
\usepackage{multirow}
\usepackage{pgfplots}
\usepackage{tikz}

\begin{document}

\title{PDANet: Pyramid Density-aware Attention Net for Accurate Crowd Counting}

\author{Saeed~Amirgholipour,~\IEEEmembership{Member,~IEEE},~
		 Xiangjian~He,~\IEEEmembership{Senior Member,~IEEE},~
		Wenjing Jia,~\IEEEmembership{Member,~IEEE}, 
		Dadong~Wang,~\IEEEmembership{Senior Member,~IEEE}, 
		~Lei~Liu,

		\thanks{Saeed Amirgholipour, Xiangjian He and Wenjing Jia are in Global Big Data Technologies Centre, University of Technology Sydney, Australia. Email: Saeed.AmirgholipourKasmani@student.uts.edu.au; Xiangjian.He@uts.edu.au; Wenjing.Jia@uts.edu.au;.}
		\thanks{Dadong~Wang is with the Quantitative Imaging Research Team, CSIRO Data61, Australia. E-mail: ~Dadong.Wang@csiro.au.}
		\thanks{ Lei Liu was in the School of Instrumentation Science and Opto-Electronics Engineering, Beihang University, China. E-mail:by1417114@buaa.edu.cn.}
\thanks{Corresponding author: Xiangjian He.}}

\IEEEtitleabstractindextext{%

\begin{abstract}
Crowd counting, \textit{i.e.}, estimating the number of people in a crowded area, has attracted much interest in the research community. Although many attempts have been reported, crowd counting remains an open real-world problem due to the vast scale variations in crowd density within an interested area, and severe occlusion in a crowd. 
In this paper, we propose a novel Pyramid Density-Aware Attention-based network, abbreviated as PDANet, which leverages the attention, pyramid scale feature and two branch decoder modules for density-aware crowd counting. 
The PDANet utilizes these modules to extract different scale features, focus on the relevant information, and suppress the misleading ones. 
We also address the variation of crowdedness levels among different images with an exclusive Density-Aware Decoder (DAD). 
For this purpose, a classifier evaluates the density level of the input features and then passes them to the corresponding high and low crowded DAD modules. 
Finally, we generate an overall density map by considering the summation of low and high crowded density maps as spatial attention. 
Meanwhile, we employ two losses to create a precise density map for the input scene. 
Extensive evaluations conducted on the challenging benchmark datasets well demonstrate the superior performance of the proposed PDANet in terms of the accuracy of counting and generated density maps over the well-known state-of-the-art approaches.
\end{abstract}

}

\maketitle

\IEEEdisplaynontitleabstractindextext

\IEEEpeerreviewmaketitle

\section{Introduction}
\label{sec:introduction}

\IEEEPARstart{N}{owadays}, crowd counting has become an important task for a variety of applications, such as traffic control~\cite{liu2019context}, public safety ~\cite{ling2019indoor}, and scene understanding ~\cite{shao2015deeply,ling2019indoor}. 
As a result, density estimation techniques have become a research trend for various counting tasks. 
These techniques utilize trained regressors to estimate people density for each area so that the summation of the resultant density functions can yield the final count of crowd. A variety of regressors, such as Gaussian Processes~\cite{chan2009bayesian}, Random Forests ~\cite{lempitsky2010learning}, and more recently, deep learning based networks~\cite{liu2019denet,li2018structured,liu2019performance} have been used for crowd counting and density estimation. 
However, the state-of-the-art approaches are mostly deep learning based approaches due to their capabilities of generating accurate density maps and producing precise crowd counting~\cite{liu2019context,zhang2019nonlinear}.

Generally, the approaches based on deep neural networks (DNNs) utilize standard convolutions and dilated convolutions at the heart of the models to learn local patterns and density maps~\cite{liu2019performance,amirgholipour2018ccnn}. 
Most of them use the same filters, pooling matrices, and settings across the whole image, and implicitly assume the same congestion level everywhere~\cite{liu2019denet}. 
However, this assumption often does not hold in reality. 

\begin{figure}
    \centering
    \includegraphics[width=6 cm]{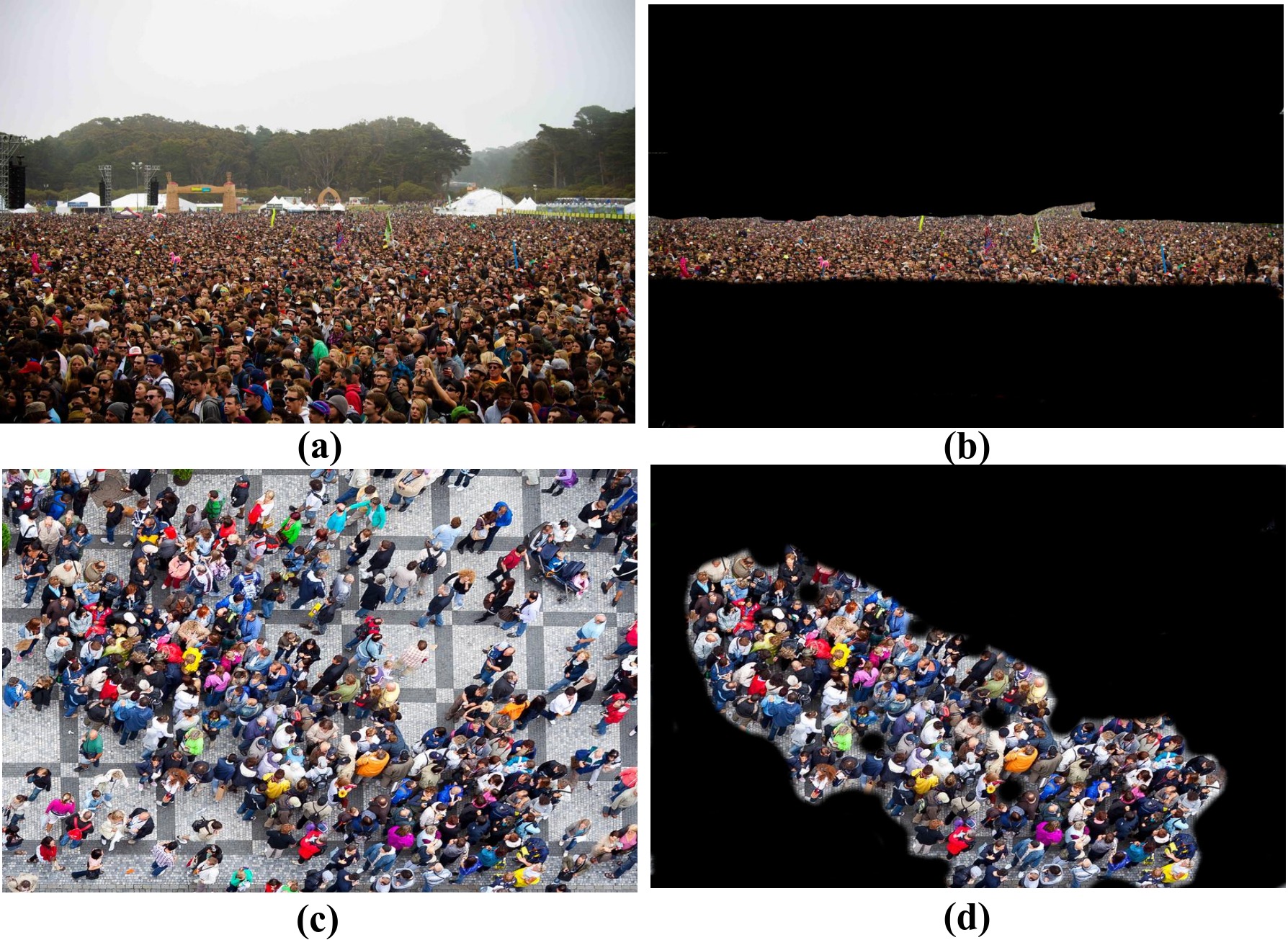}
    \caption{Examples of crowded and sparse images. (a) and (c) show an example of a highly crowded scene and a less crowded scene, respectively, while (b) and (d) show their corresponding congested areas. }
    \label{fig:fig1}
\end{figure}

To better understand the effect of this mis-assumption, let us show some examples with clearly different levels of crowdedness. 
Fig.~\ref{fig:fig1} presents some exemplar images of different congestion scenarios. 
Fig.~\ref{fig:fig1}(a) shows a highly crowded image having more than 1,000 people, while Fig.~\ref{fig:fig1}(c) presents a less crowded scene having less than 70 people. 
However, if we look at Fig.~\ref{fig:fig1}(a), we notice that there is a relatively more congested area, which is shown in Fig.~\ref{fig:fig1}(b). The same situation can be seen in Fig.~\ref{fig:fig1}(c), and it is obvious that a small area within this crowd, as shown in Fig.~\ref{fig:fig1}(d), is more crowed.
Due to this dynamic variation in the crowded scenes, naturally we should
utilize different features and branches to respond and capture details at different levels of crowdedness. 
In the past, this has been attempted by four major types of approaches, \textit{i.e.}, defining separate pathways from the
lower layers and utilizing different sizes of the convolutional filters, image pyramid-based methods~\cite{chen2019spn,liu2019context}, detection-based crowd counting~\cite{liu2019denet}, patch-based crowd counting~\cite{sam2017switching,amirgholipour2018ccnn,sam2019almost}, and multi-level feature based methods~\cite{chen2019spn}.
Although these methods achieved robust performance with some different tactics, there are still lots of spaces to improve their performances by designing highly efficient convolutional layer structures, which can effectively deal with crowd scenes with dramatic density varieties effectively. 

First, generally speaking, a kernel size of \( 3\times 3\) for a convolution filter is more effective than the larger ones in terms of extracting more meaningful features, because more details can be captured with lower complexities without making it more difficult to train the network~\cite{simonyan2014very,szegedy2015going,szegedy2016rethinking}. Kang \textit{et al.}~\cite{kang2018crowd}
proved that smaller receptive fields gave better performance. 
Secondly, using patch-based processing and multi-patch processing  is time costly due to that the same features have to pass through different paths and patches multiple times. 
If we want to take the benefits of multi-patch or multi-column based approaches, it is better to extract some coarse features from the initial layers and then pass them to some branches for further zooming in to find more sophisticated features. To utilize a deeper network for crowd counting, we need an approach that can deploy the aforementioned proposals on the multi-column structure to achieve better performance. 

In this paper, we present a deep encoder-decoder based architecture named as Pyramid Density-aware Attention-based Network (PDANet), which combines the pyramid feature extraction with spatial and channel attentions to produce richer features estimating crowd of various levels of crowdedness and scales. 
In our work, we use the VGG16 as the feature extractor for the encoder to produce features for the decoder of the model. 
To learn multi-scale features, we first use a
cascade of Global Average Pooling (GAP), \( 1\times 1\)  convolution and dilated convolutions with kernels of $3 \times 3$ to extract more mature features with different scales from VGG16 features. 
Then, we apply the channel and spatial attentions in different layers to enhance and boost the quality of features in order to obtain more accurate density maps. 
On the other hand, to make the model adaptive to different density levels within an image, we introduce a classification module to
classify the crowdedness level of the input scene and develop generation models of  low and high crowded density maps for the input image.

This work is different from the existing crowd counting approaches that use the pyramid contextual information and attention modules in several ways. Unlike our previous model, DENet [5], the propose PDANet does not separate models for counting people in sparsely crowed areas and estimating the human density maps in the remaining areas in an image. (b) The first main characteristic of our proposed PDANet is its density awareness by adopting the pyramid and attention modules. Different from other works attempting to address this problem of density variety, \textit{e.g.}~\cite{sam2017switching}, our PDANet does not separate the input scene into different patches. Instead, we use multipath branching to address the intra-density variations within the input scene. Experimental results show that, the pyramid and attention modules contribute a 5 to 20 percent improvement over the baseline model. (c) Pyramid Feature Extractor (PFE) is the second noticeable contribution of our PDANet. We utilize a new combination of GAP, \( 1\times 1\) convolution, and Atrous convolution, resulting in a difference from the existing... approaches in terms of the orders and parameters that can better aggregate local scale features and is more effective than the existing solutions. (d) The third remarkable feature of PDANet is its attention modules. The architecture of our end-to-end attention modules is also different from ADCrowdNet~\cite{liu2019adcrowdnet} because it uses the combination of the spatial and channel based attention modules within the architecture. Furthermore, it is trained in an end-to-end way based on the crowd counting dataset, instead of separately using external dataset to train the attention module as in ADCrowdNet~\cite{liu2019adcrowdnet}. Compared with the work in~\cite{wu2019adaptive}, our PDANet has also adopted another spatial-based module in the DAD module to optimize the density map results based on feature maps of the sparse and dense areas within the input scene. (e) The last distinct characteristic of the PDANet is classification modules, which are different from the existing work~\cite{wu2019adaptive}. Our PDANet passes the input image to two different sub-models with different receptive fields to evaluate lower and higher bounds of the density map, and then combines them with the help of channel attention module. Our PDANet introduces a classification module that classifies the input image to the low or high-density data and passes them to the corresponding and appropriate DAD modules. 

To summarize, the contributions made in this paper are as follows.
\begin{itemize}

\item In order to address crowd areas of various scales and density levels, we propose a density-aware solution, which is achieved with the combination of multi-scale feature extraction, density classification and adaptive density estimation modules. 
This feature helps the model to handle density variation between different images as well as within each input scene. 

\item We first integrate the pyramid multi-scale feature extraction mechanism in a feature extractor to extract rich features for the following classification module. Then, we integrate the channel and spatial attention modules and propose an end-to-end trainable density estimation pipeline. Both modules contribute to exploit the right context at each location within a scene. 

\item For estimating densities of crowd with not only high and low crowdedness levels but also inter-level density areas, we propose to use a combination of classification and regression losses to address the whole and within-the-scene changes in the density maps.

\item Extensive experiments on several challenging benchmark datasets are conducted to demonstrate the superior performance of our proposed PDANet approach over the state-of-the-art solutions. We also preform comprehensive ablation studies at Section   \text{II} of the supplementary document of this submission to validate  the effectiveness of each component in our proposed approach.
%
\end{itemize}
The rest of the paper is organized as follows. In Section \(\text{II}\), we introduce the existing works related to our approach. The proposed PDANet model for crowd density estimation is introduced in detail in Section \(\text{III}\). In Section  \(\text{IV}\), we evaluate the performance of PDANet on benchmark datasets. Finaly, we draw conclusions in Section  \(\text{V}\).

\section{Related Works}
\label{sec:related}
In this section, we provide literature review related to our PDANet model. Many studies have been done based on multi-column architectures~\cite{sindagi2019ha,tian2019padnet}. 
One of the initial works was done by Zhang \textit{et al.}~\cite{zhang2016single}, who proposed a three-CNN-column based MCNN structure, each with different receptive
parameters to handle a range of different head sizes. 
Recently, Tian \textit{et al.} proposed PaDNet~\cite{tian2019padnet}, which was composed of several components such as  the Density-Aware Network (DAN), Feature Enhancement Layer (FEL), and  a Feature Fusion Network (FFN). PaDNet improved the-state-of-the-art results remarkably by capturing pan-density information and utilizing global and local contextual features. IG-CNN~\cite{babu2018divide} was another extensive study that combined the clustering and crowd counting for estimating the density map more adaptively based on training a mixture of experts that could incrementally adapt and grow based on the complexity of the
dataset. 
Sindagi \textit{et al.} proposed a new multi-column network, \textit{i.e.}, CP-CNN~\cite{sindagi2017generating}, which added two other branches to classify an image-wise density to provide the global and local context information to the MCNN model. 
Deb \textit{et al.}~\cite{deb2018aggregated} incorporated the Atrous convolutions into the multi-branch network by assigning different dilation rates to various branches. 

Most recently, Shi \textit{et al.}~\cite{shi2019revisiting} proposed a perspective information CNN-based model PACNN for crowd counting. 
Their model combined the perspective information with a density regression to address the person scale change within an image. 
They generated the ground truth of perspective map and used it for generating perspective-aware weighting layers to combine the results of multi-scale density adaptively. 
Wan \textit{et al.}~\cite{wan2019residual} proposed a new model RRSP to utilize the correlation information in a training dataset (residual information) for accurate crowd counting. 
They fused all the residual predictions and created the final density map based on the appearance-based map and the combination of residual maps from the input scene. 

Some of the recent studies focused on utilizing pyramid and attention-based modules~\cite{varior2019scale}. 
Pyramid modules were introduced by Zhao \textit{et al.}~\cite{zhao2017pyramid} to produce proper quality features on the scene semantic segmentation task. 
They introduced an efficient method to estimate human head sizes and integrated them to an attention module to
aggregate density maps from different layers and generate the final density map. 
Liu \textit{et al.}~\cite{liu2019context} presented another end-to-end multi-scaled solution CAN based on fusing multi-scale pyramid features. They used modified PSP modules for extracting multi-scale features from the VGG16 features to address the rapid scale changes within the scenes. Their model leveraged multi-scale adaptive pooling operations to
cover a variety range of receptive fields. 
Compared to CAN, Chen \textit{et al.} proposed an end-to-end single-column structure as a Scale Pyramid Network (SPN), which extracted multi-scale features with the dilated
convolutions with various dilation rates (\( 2,4,8, \) and \( 12 \)) from the VGG16 backbone features~\cite{chen2019spn}. The experimental results proved that their idea worked well on some well-known datasets. 

On the other hand, the attention module and idea proposed in~\cite{hu2018squeeze} aimed to re-calibrate the features adaptively, so as to highlight the effect of valuable features, while suppressing
the impact of weak ones~\cite{roy2018concurrent}. 
Recently, researchers attempted to incorporate this module and its variations into their
models to improve the performance in several tasks such as object detection, object classification, and medical image processing~\cite{jetley2018learn, schlemper2018attention,zagoruyko2016paying}. 
Rahul \textit{et al.} proposed an attention-based model to
regress multi-scale density maps from several intermediate layers~\cite{liu2019adcrowdnet}. We recently proposed DENet ~\cite{liu2019denet} base on a combination of crowd counting and density estimation.  DENet utilized mask-RCNN for counting people in the low crowded areas and the Xception based regressor for regressing people in the highly crowded areas. We proved that DENet could achieve excellent results in the low crowded images as well as highly crowded images.
ADCrowdNet~\cite{liu2019adcrowdnet} was one of the latest research in the area of crowd counting, and it used attention modules to generate accurate density maps. 
Liu \textit{et al.} utilized a two-step cascade encoder-decoder architecture, one for the detection of crowded areas and producing the attention map as Attention Map Generator (AMG), and the other for generating the density map called Density
Map Estimator (DME).  Their method achieved excellent results on the ShanghaiTech Part A dataset. Their method achieved excellent results on the ShanghaiTech Part A dataset. Although the idea of using the attention map was interesting, it has some significant drawbacks, such as that (a) it needed an external dataset to train AMG to detect the crowd area, and (b) after producing the attention map, they applied it on the input scene to create masked input data
for DME and again extracting features with a similar encoder-decoder structure. We believe that it is redundant and time consuming due to passing the input scene twice rather than appling the generated mask on the latest layer of AMG module and use the feature maps for the next stage.

\section{Pyramid Density-aware Attention Net}
\label{sec:PDANet}

In this section, we first present the general structure of our proposed PDANet for adaptively addressing the challenges in crowd counting. 
This new structure uses pyramid-scale feature extraction and consists of adaptive pooling, and \( 1\times 1\) and \( 3\times 3\) convolutions to enrich the feature maps for handling objects of various scales within a scene. 
In the following subsections, we will give more details about the attention modules, pyramid feature modules, Classification Module and decoders. The loss functions that this paper uses are defined in Section  \text{I} of the supplementary document.

\begin{figure}[t]
    \centering
    \includegraphics[width=0.45\textwidth]{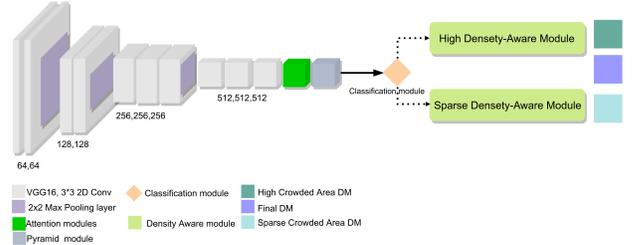}
    \caption{The overview of our proposed PDANet network. This architecture contains a VGG16 based feature extractor, a Pyramid module, an Attention module, a Classification module, and a Decoder module.}
    \label{fig:fig2}
\end{figure}

\subsection{ Overview}
\label{sec:overview}

As discussed above, we formulate crowd counting as the problem of regressing people's density map from a scene. 
The overall architecture of our PDANet for regressing the density map of the crowd from an image is illustrated in Fig.~\ref{fig:fig2}. This framework contains five main components, \textit{i.e.}, a Feature Extractor, a Pyramid Feature Extractor (PFE), a Classifier, a Density Aware Decoder (DAD), and an Attention Module. 
Each of these components contributes to the overall accuracy and efficiency of the model for crowd counting.

The backbone of our PDANet is a network based on VGG16~\cite{simonyan2014very}, which is widely used for extracting low-level features. 
We eliminate the layers between the last two pooling layers considering the trade-off between resource cost and accuracy~\cite{li2018csrnet}. 
Then, we apply a channel and spatial based attention module to it to highlight essential features. 
Then, these features are fed into the PFE module, which incorporates the combination of adaptive pooling and \( 1\times 1\) and \( 3\times 3\) dilated convolutions to produce scale-aware mature features for last layers of the decoder module. 
In the next step, we incorporate a GAP and a fully connected layer to classify the input scene as a highly dense or a sparse one. 
Then, we pass this information to the respective decoder with the same structure (our theoretical studies proves that the same respective field is better than a different one). 
The decoder contains four \( 3\times 3\) dilated convolution layers, which are empowered with an attention module after each layer. 
Furthermore, to address the congestion differences in sparse and dense areas, we design two branches of the decoder module to generate low and high-density maps within the input scene and assign them to the corresponding regression losses. 
In the final step, we use the dense and sparse features from the last layer of the decoder to produce the final output density map (DM). 
Our PDANet uses the same loss for sparse, dense and final output DMs, and a classification loss to train the model in an end-to-end manner.

To summarize, in our proposed PDANet, each part plays a role in the overall performance.
\begin{itemize}
\item The Attention Module focus its attention on the significant features (crowded areas). 
\item The Pyramid Feature Extractor generates more productive features, which are more suitable for the crowd counting task with scale variation, through a combination of adaptive pooling algorithms and dilated convolutions with different scales.
\item  The Classifier helps find the proper branch of the decoder according to the crowdedness level of the area.
\item The mid-branch Decoder is to address congestion change within the input image.
\end{itemize}

\subsection{Channel and Spatial based Attention Modules}
\label{sec:am}

\begin{figure}

    \centering
    \includegraphics[width=6 cm]{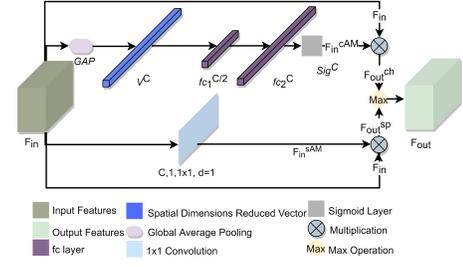}
    \caption{Illustration of the attention module of our model. The top branch generates channel-based attention, while the bottom branch generates the spatial attention map.}

    \label{fig:fig3}
\end{figure}

In this study, we re-calibrate the feature maps adaptively by mixing attention modules to augment the effect of essential features, while suppressing the weak ones. 
We use the combination of spatial and channel-based attentions for finding and separating the crowded areas within the input image. 
As it is shown in Fig.~\ref{fig:fig2}, we utilize an attention module in our model, which is the channel and spatial attention~\cite{roy2018concurrent} after the convolution layers, shown as the green module in Fig.~\ref{fig:fig2}. 
This module contains the channel and the spatial attentions to produce the final attention features in each layer. We combine the results of these two attentions by an element-wise max of the channel and the spatial excitations to generate output features in each layer. The other attention module is a spatial attention map that is generated based on the density map of the sparse and dense crowded areas within the image. We apply a sigmoid on this attention module and multiply it with the joint convolution feature maps from the last layer of a sparse and dense decoder. 

\begin{figure*}[]
    \centering
    \includegraphics[width=0.7\textwidth]{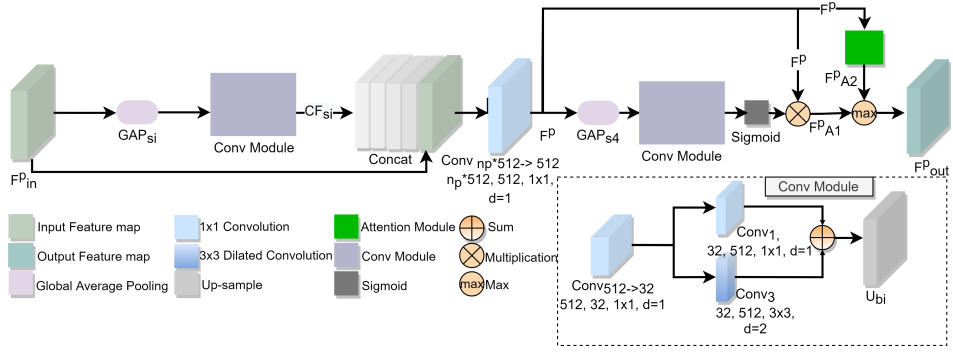}
    \caption{The overview of the Pyramid Feature Extractor (PFE) module. The PFE module uses \( 1\times 1\) and \( 3\times 3\) dilated kernel convolutions with the GAP to extract features of different scales from the VGG16 features.}
    \label{fig:fig4}
\end{figure*}

Fig.~\ref{fig:fig3} illustrates this attention module. 
As shown in this figure, there are two branches in this illustration, \textit {i.e.}, the channel attention branch on the top, and the spatial attention branch on the bottom. 
The channel attention branch utilizes a cascade of GAP and two fully connected layers with the size of \(\frac{C}{2}\) and \(\emph{C}\), respectively (\(\emph{C}\) is the channel size of a convolution layer). Then, after applying a sigmoid on the result, we do element-wise multiplication between the channel attention map and the input feature maps to obtain channel-wise weight corrected feature maps.

As we explained in the previously, to apply the channel based attention mechanism, we first perform GAP on the input feature map \(F_{in}\), to obtain \(V^{C}\), and then transform them by two fully-connected layers \( {f_c}_{1}^{C/2}\) and  \({f_c}_{2}^{C}\),  as shown in Fig.~\ref{fig:fig3} and Eq.~\ref{eq:1} as:
\begin{equation}
\small
F_{in}^{cAM} = \text{Sig}( {f_c}_{2}^{C}({f_c}_{1}^{C/2} (V^{C}))).
\label{eq:1}
\end{equation}
where \({\text{Sig}} \)  is a sigmoid function that yields the value in a range of \( [0, 1]\) to find the impact of each layer in the feature maps. 

Therefor, for channel based attention, features \(F_{out}^{ch}\) are obtained by multiplying the encoded channel-wise dependencies (  \(F_{in}^{cAM}\)  ) with \(F_{in}\) to get \(F_{out}^{ch}\). On the other hand, to obtain the spatial attention map, \(F_{in}^{sAM}\), we perform a \( 1\times 1\) convolution, \textit{i.e.}, \(Conv \in \mathbb{R}^{1 \times C\times 1\times 1}\),  on the input feature map.  
Thus, we can measure the importance of a spatial information of each pixel or location within  \(F_{in}\). 
In the next stage, we multiply the spatial attention map with the input feature maps to get the final spatial attention features \(F_{out}^{sp}\), which augment relevant spatial locations and suppress irrelevant ones. Finally, we combine the results of these two attentions by element-wise max of the channel and spatial excitation, \textit{i.e.}, \(F_{out} = \text{max}(F_{out}^{ch}, F_{out}^{sp}) \). These feature maps amplify the input feature map data and re-calibrate the crowded area within each input convolution layer.

\subsection{Pyramid Feature Extractor (PFE)}
\label{sec:pfe}

In this section, we propose a Pyramid Feature Extractor (PFE), which is inspired by the Spatial Pyramid Pooling~\cite{chen2019spn} to address this issue. 
The PFE fuses features under various pyramid scales by a combination of GAP and two shared 2D convolution layers with a mixture of \( 1\times 1\) and \( 3\times 3\) dilated kernels. 
The general operation of PFE is illustrated in Fig.~\ref{fig:fig4}. 

We extract contextual features by various GAP. In PFE module, we keep the ratio of the input feature map with \( \text{GAP}_{s_i}\) at scale \( s_i\), for $i=2, 3, \cdots,10$ and produce contextual features for each channel with a size of \( H{s_i} \times W_{s_i} \). For example, if we have an input feature map with  \(  \mathbb{R}^{1 \times C\times H\times W}\), \( \text{GAP}_{s_2}\) utilize global averge pooling layer to generate scaled feature map with size of  \(  \mathbb{R}^{1 \times C\times {H \over 2} \times {W \over 2}}\), where  \( H_{s_2} \) and \( W_{s_2} \) are equal to \( H \over 2\) and \( W \over 2 \), respectively. Various scales of contextual features form the pooled representations for different areas and provide rich information about the density levels in various sub-regions of the input image.  The results presented in the Experiments section are based on the scenario utilizing three \( \text{GAP}_{s_i}\) with scale of  \(s_2\) ,  \(s_4\) , and  \(s_8\), respectively. In the Ablation Study shown in Section \text{II} of the supplementary document, we  compare several scenarios for the use of \( \text{GAP}_{s_i}\). 

Then, we feed \( \text{GAP}_{s_i}\) to the Conv Module to improve the representation power of the feature map. 
This procedure is different from the architectures that reduce the dimension of the input feature map with convolution~\cite{chen2019spn}. As illustrated in Fig.~\ref{fig:fig4}, we perform the Conv operation as:
\begin{multline}
\small
\label{eqn:pm2}
\text{CF}_{s_i} = U_{bi} (\text{Conv}_{1}(\text{Conv}_{512 \to 32}(\text{GAP}_{s_i})) \\ 
+  \text{Conv}_{3}(\text{Conv}_{512 \to 32}(\text{GAP}_{s_i}))).
\end{multline}
where, for each scale \(s_i\), \( \text{CF}_{s_i}\), is the shared Conv module that comes with a bi-linear interpolation to up-sample the contextual features \( U_{bi}\) to the ones of the same size as that of \( F_{in}^{p} \). 

The shared layer contains one \( 1\times 1\) convolution (\(\text{Conv}_{512 \to 32}\)) to reduce the number of channels from 512 to 32. 
We do this to reduce the number of parameters that need to train and reduce the computational cost of PFE. In the subsequent stage, we get the summation of a \( 1\times 1\) convolution (\(\text{Conv}_{1}\)), and a \( 3\times 3\) dilated convolution (\(\text{Conv}_{3}\)) as a piece of extra bonus information that captured from surounding contextual features within \( \text{GAP}_{s_i}\). Experimentally, we verify that this combination of convolution filters improves the performance of the PFE module in the density estimation task. Finally, we concatenate all  \( \text{CF}_{s_i}\) and the input features \( F_{in}^{p}\) with a \( 1\times 1\) convolution. 
We reduce the number of the channels to the original VGG features \( F_{in}^{p}\). We define this as:
\begin{equation}
\label{eqn:pm3}
F^{p} = \text{Conv}_{n_p*512 \to 512} (\text{Concat}(\text{CF}_{s_i},F_{in}^{p})). 
\end{equation}
where \( n_p \) is the number of pyramid contextual features \( \text{CF}_{s_i}\), plus  the features in the original feature map, and  \(  \text{Conv}_{n_p*512 \to 512} \) is a \( 1 \time 1 \) convolution to reduce the number of channel to 512. 

Then, we utilize a special attention module, which is the combination of the Conv module and attention module that we explained in Section~\ref{sec:am}. 
We pass \( F^{p}\) to two separate attention branches. 
As illustrated in Fig.~\ref{fig:fig4}, in the bottom, we feed the \( F^{p}\) to the  \( \text{GAP}_{s_4}\) layer and reduce the input size to  ${H\over 4}\times {W\over 4}$, and then apply the Conv module to it. We apply the GAP to highlight and escalate the most important parts of the output feature maps. Then, after performing \(\text{ Sig}\) on the output of conv module, we apply by the element-wise multiplication to produce the attention map by Conv module (\(F_{A1}^{p} \)). On the top, we also perform the attention module that we discussed in the session ~\ref{sec:am} to generate the attention feature map output (\(F_{A2}^{p} \)). Finally, we combine the results of these two attentions by the element-wise \(\text{ max} \) operation of the Conv module output (\(F_{A1}^{p} \)) and the attention module output (\(F_{A2}^{p} \)) as defined by:
\begin{equation}
\small
\label{eqn:pm4}
F_{out}^{p} = \text{max}(F_{A1}^{p},F_{A2}^{p}).
\end{equation}

Altogether, as illustrated in Fig.~\ref{fig:fig4}, the PFE module extracts contextual features \( \text{CF}_{s_i}\) as discussed above, and then feed them to the classification module and a Density Aware Decoder (DAD) module that produces the density map.

\subsection{Classification Module}
\label{sec:classification}

\begin{figure}
    \centering
    \includegraphics[width=5 cm]{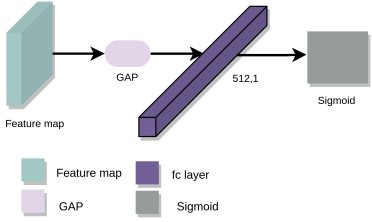}
    \caption{Illustration of the classification module of PDANet. It uses the global  average pooling with a fully connected layer to determine the dense level of input scene.}
    \label{fig:fig5}
\end{figure}

The next step in our overall framework, as illustrated in Fig.~\ref{fig:fig2}, is to decide whether the input contextual features are dense or sparse. 
We do this to address the huge variation of crowd densities among different images. 
We pass input features to the suitable DAD to adaptively react to the density level of the input image and provide a better estimation for crowd density. 
To model this, as shown in Fig.~\ref{fig:fig5}, we introduce a binary classification module to learn how to classify the input feature maps into two classes, \textit{i.e.}, dense or non-dense (\textit{aka}, sparse), as:
\begin{equation}
\small
\label{eqn:cl}
\text{Cl}_{t}^{est} = \text{ Sig}(f_c(\text{GAP})).
\end{equation}
where \(\text{GAP}\) is global average pooling with the scale of \( 1\times 1\) and produces a vector with the size of \(512\), \(f_c\) is a fully connected layer, and  \( \text{ Sig} \) is the sigmoid function that yields the value in a range of [0,1] to find the impact of each layer in the feature maps. 

Thus, the classification module produces a class probability, which is a value in the range of [0,1]. If the output probability (\(\text{Cl}_{t}^{est}\)) is less than \( 0.5 \), the model considers the input as a non-dense crowd image and passes it to the sparse DAD branch. Otherwise,  it passes it to the high DAD branch, as shown in Fig.~\ref{fig:fig2}.

\subsection{Density Aware Decoder (DAD)}
\label{sec:dad}

DAD is one of the special modules of our proposed PDANet model, as it dynamically handles intra-variation of the density level within the input image. 
To achieve this, we use four dilated convolution layers with the attention module attached to each layer, similar to the one introduced in Section~\ref{sec:am}. 
According to the result of the classification module, we pass the output of the PFE module (\(F_{out}^{p}\)) to one of two DAD modules. If the input scene is highly crowded, we direct \(F_{out}^{p}\) to the high DAD branch. Otherwise, we pass it to the sparse branch. We achieve a model that can address the density variation of among different input image adaptively. Furthermore, the DAD module by itself is composed of two parts, \textit{i.e.}, the shared layers, and the low or high-density decoder branches. This design enables us to cope with various occlusions, internal changes, and diversified crowd distributions within every single input scene, as illustrated in Fig.~\ref{fig:fig1}.

\begin{figure}[t]
    \centering
    \includegraphics[width=0.4\textwidth]{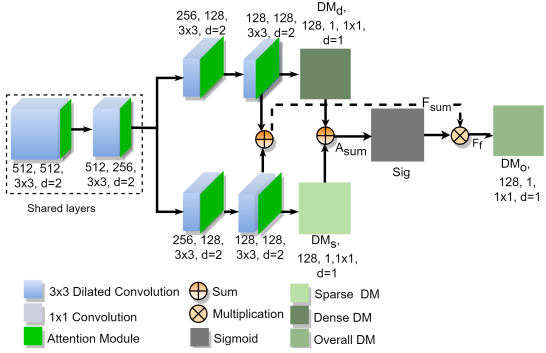}
    \caption{The illustration of the DAD module. The input feature maps are fed to the two shared layers and then we use the two branches with three convolution layers to handle the dense and sparse areas within the scene.}
    \label{fig:fig6}
\end{figure}

The structure of DAD is illustrated in Fig.~\ref{fig:fig6}. As shown in the figure, we consider the first two layers as shared layers and then pass the output feature map along two separate paths with the other three convolution layers to manage the within-image density variation, as shown in Fig.~\ref{fig:fig1}. The number of channels in the dilated convolution in DAD is (\(N_{ch} = {512,256,128,128,1}\)) with the kernel filter size \( 3\times 3\) and the dilation rate \(d_{rate} = 2\) for the first four layers and \( 1\times 1\) convolution at the end to produce the density maps. We call the outpot of dense and sparse branches as \( \text {DM}_{d} \) and \( \text {DM}_{s}\), respectively.
Furthermore, to reduce the number of training parameters, we utilize a \( 1\times 1\) convolution to reduce the input channels to \( 32\) and then perform a 2D dilated convolution on the reduced channel feature maps. This process speeds up the training and convergence of our model.

Moreover, there is a small notation for the dense and non-dense crowded areas. We use the \(\text {DM}_{d}\) for the high dense regions within the image. However, for the low density regions, within a low or highly dense input image, we use a shared \( \text {DM}_{s}\) layer. 
This design gives us the benefit of using more information to train the model to map the low and dense regions with the input image. 
Therefore, we are able to have a better density estimation for the low crowded areas. 
On the other hand, by utilizing a different \(\text {DM}_{d}\) for the highly dense areas within the input image, our DAD module is able to improve its estimation for these areas too.

By utilizing this architecture in the DAD, we will have two resultant density maps for the low and high crowded areas of the input image. Besides this, we pick up these feature maps of the last layer in the dense and non-dense branches. 
Then, we sum up these feature maps to form an attention module \(A_{\text{sum}}\), and name the summation as  \(F_{\text{sum}}\). 
Therefore, we use the following equation to produce the final overall feature map:
\begin{equation}
\small
\label{eqn:attL}
F_{f} = F_{sum} \times \text{ Sig}(A_{\text{sum}}),
\end{equation}
where \(\text{ Sig}(A_{\text{sum}})\) is the sigmoid scaling of \(A_{\text{sum}}\), and \(F_{f}\) is the final overall feature map, which is fed to the final layer to produce an overall dense map.

\section{Experiments}
\label{sec:exp}

In this section, we evaluate the performance of our proposed approach. 
We first introduce the evaluation metrics and then report experimental results obtained on benchmark datasets. The experiments are conducted on four benchmark datasets, and results are compared with the recently published state-of-the-art approaches, which have already been used for comparison purpose since. The detailed ablation study is also made in
Section \text{II} of the supplementary document. 

\subsection{Evaluation Metrics}
\label{sec:em}

Previous works on crowd density estimation used the Mean
Absolute Error (\(\text{MAE}\)) and the Root Mean Squared Error (\(\text{MSE}\)) as evaluation metrics~\cite{cao2018scale,li2018csrnet,chen2019spn,liu2019context}, which are defined by: 

\begin{equation}
\small
\text{MAE}=\dfrac{1}{M}\sum_{m=1}^{M} \left | C_{m}^{est}- C_{m}^{gt} \right |.
\label{eq:n1}
\end{equation}
and
\begin{equation}
\small
\text{MSE}=\sqrt{\dfrac{1}{M}\sum_{m=1}^{M} \left ( C_{m}^{est}- C_{m}^{gt} \right )^{2}}.
\label{eq:n2}
\end{equation}
where \( M \) is the number of testing images, \(C_{m}^{gt}\) denotes the exact number of people inside the ROI of the \( m \)-th image and \( C_{m}^{est}\) is the correspondingly estimated number of people. 
In the benchmark datasets discussed below, the ROI is the
whole image except when explicitly stated otherwise. We follow the methodology in ~\cite{cao2018scale} to prepare ground truth density data. Note that the number of people in an image can be calculated by the summation over the pixels of the ground truth  (\(D_{i}^{gt}  \)) as it is defined in Eq. $2$  in the supplementary document and the predicted density maps (\(  D_{i}^{est} \)).

\setlength{\tabcolsep}{7pt}
\begin{table*}
\scriptsize
\begin{center}
\caption{Comparison of the MAE and MSE results obtained with our proposed PDANet and the-state-of-the-art crowd counting approaches on the ShanghaiTech Part A and Part B~\cite{zhang2016single}, the WorldExpo10 ~\cite{zhang2015cross}, and the UCF CC 50~\cite{idrees2013multi} as high crowded dataset and  UCSD as low crowded dataset~\cite{chan2008privacy}.}
\label{table:1}
\begin{tabular}{l|ll|ll|llllll|ll|ll}

\hline\noalign{\smallskip}
 $\qquad\qquad$&\multicolumn{2}{c|}{ShanghaiTech A} &\multicolumn{2}{c|}{ShanghaiTech B} & \multicolumn{6}{c|}{WorldExpo10}&\multicolumn{2}{c|}{UCF CC 50}& \multicolumn{2}{c}{UCSD} \\
\hline\noalign{\smallskip}
Methods $\qquad\qquad$& \text{MAE} & \text{MSE} & \text{MAE} & MSE& Sce.1 & Sce.2& Sce.3& Sce.4& Sce.5& AVG& \text{MAE} & MSE& \text{MAE} & \text{MSE} \\
\noalign{\smallskip}
\hline
\noalign{\smallskip}
   
 A-CCNN~\cite{amirgholipour2018ccnn}&85.4	&124.6&19.2	&31.5	&-	&-&-	&-	&-	&-	&367.3	&423.7& 1.36 & 1.51 \\
BSAD~\cite{huang2017body}&90.4	&135.0&20.2	&35.6&4.1	&21.7	&11.9&	11.0	&3.5	&10.5&409.5&	563.7& 1.00	&1.40\\
ACSCP~\cite{shen2018crowd}  & 75.7 & 102.7 & 17.2 & 27.4 & 2.8	&14.05	&9.6&	8.1	&2.9	&7.5& 291.0 & 404.6 & 1.04	 &1.35\\
D-ConvNet-v1~\cite{zhang2019nonlinear}  & 73.5 & 112.3 & 18.7 & 26.0 & 1.9	&12.1&	20.7	&8.3	&2.6	&9.1 & 288.4 & 404.7&-	&-  \\
IG-CNN~\cite{babu2018divide}  & 72.5 & 118.2 & 13.6 & 21.1 & 2.6	&16.1	&10.15	&20.2	&7.6	&11.3  & 291.4 & 349.4&-	&-  \\
DRSAN~\cite{liu2018crowd}  & 69.3 & 96.4 & 11.1& 18.2 & 2.6		&11.8	&10.3	&10.4	&3.7	&7.76 & 219.2& 250.2 &-	&- \\
ic-CNN~\cite{ranjan2018iterative}  & 68.5 & 116.2 & 10.7 & 16.0 & 17.0	&12.3	&9.2	&8.1	&4.7	&10.3   & 260.9& 365.5&-	&- \\

CSRNet~\cite{li2018csrnet}  & 68.2 & 115.0 & 10.6 & 16.0 & 2.9	&11.5	&8.6	&16.6	&3.4&	8.6& 266.1 & 397.5  &1.16	 &1.47\\
SANet~\cite{cao2018scale}  & 67.0 & 104.5 & 8.4 & 13.6& 2.6	&13.2	&9.0	&13.3&	3.0&	8.2 & 258.4 & 334.9 &  1.02 &	1.29\\

SFCN~\cite{wang2019learning}  & 64.8 & 107.5 &  7.6 & 13.0 &-	&-	&-	&-	&-	&- & 214.2 & 318.2&-	&-  \\
TEDnet~\cite{fiaschi2012learning} & 64.2 & 109.1& 8.2 & 12.8 & 2.3	&10.1	&11.3	&13.8	&2.6	&8.0 & 249.4 & 354.5&-	&- \\

HA-CCN~\cite{sindagi2019ha} & 62.9 &  94.9& 8.1 & 13.4& -	&-	&- &-	&-	&- & 256.2 & 348.4&-	&- \\
SPN~\cite{chen2019spn} & 61.7 & 99.5 & 9.4 & 14.4 & -	&-	&-&	-	&-	&- &259.2&	335.9 &1.03	 &1.32\\
 SPN+L2SM~\cite{chen2019spn} &64.2 &98.4&7.2 &11.1 & -	&-	&-&	-	&-	&- & 188.4 & 315.3  &1.03	 &1.32\\  
ADCrowdNet~\cite{liu2019adcrowdnet} & 63.2 & 98.9& 7.7 & 12.9& \bf 1.7	&14.4	&11.5	&7.9	&3.0	&7.7 & 257.1 & 363.5 &0.98	 &1.25  \\
PACNN+CSRNet~\cite{shi2019revisiting} &  62.4 & 102.0 &   8.9 & 13.5 &  2.3&	12.5	&9.1	&11.2	&3.8	&7.8 &  267.9 & 357.8 & {\bf 0.89}	 &{\bf 1.18}\\
CAN~\cite{liu2019context} & 62.3 & 100.0 &7.8 &  12.2 &2.9	&12.0	&10.0&	7.9	&4.3	&7.4  &212.2 & 243.7&-	&- \\
DENet~\cite{liu2019denet}  & 65.5 & 101.2 & 9.6 & 15.4& 2.8	&10.7	&\bf 8.6	&15.2 &	3.5&	8.2 & 241.9 & 345.4 &1.05	&1.31\\

PaDNet~\cite{tian2019padnet}&59.2&	98.1&8.1	&12.2 & -	&-	&-&	-	&-	&-& 185.8&	278.3&0.85&	1.06\\
\midrule
 {\bf PDANet} & {\bf 58.5} &   {\bf 93.4 }& {\bf 7.1} &  {\bf 10.9}& {1.8} &  {\bf 9.1}  &9.6 &	{\bf7.3}	& {\bf2.2}	& {\bf 6.0} &  {\bf 119.8} &  {\bf 159}&  { 0.93} &  {1.21}\\
\hline
\end{tabular}
\end{center}
\end{table*}
\subsection{Data Augmentation}
\label{sec:da}
We take the benefit of data augmentation to avoid the risk of over-fitting to the small number of training images. 
We use five types of cropping alongside with a resizing as data augmentations. We crop each image into $1\over4$ of the original dimension. The first four cropped images extract four non-overlapping patches based on each corner of the original image. 
Furthermore, the fifth crop is randomly cropped from the input scene. 
For resizing, we just resize the input image to the dimension of (\(768,1024 \)) or (\(1024,768 \)) depending on the scale of the input data. If the height of an input image is bigger than the width of it, we just select (\(1024,768 \)), and in other case we resize it to (\(768,1024 \))  size.

\subsection{Experimental Results on the ShanghaiTech Dataset}
\label{sec:esh}
The ShanghaiTech dataset~\cite{zhang2016single} is one of the most popular and large-scale crowd counting datasets, and it contains 1,198 annotated images with a total of 330,165 people. It contains two parts, \textit{i.e.}, Part A (ShanghaiTech-A) with 482 images randomly collected from the Internet, and Part B (ShanghaiTech-B), including 716 images taken from the urban areas in Shanghai. As the challenge caused by diversity of scenarios and variation of congestion differs, it is difficult to estimate the number of pedestrians precisely.
  Following~\cite{cao2018scale} and as mentioned in Section~\ref{sec:em}, for setting \(\sigma \) for Part A, we use the KNN method to calculate the average distance between each head and its three nearest heads and \(\beta \) is set to \(0.3 \). 
For Part B, we set a fixed value \(15 \)  for \(\sigma \). We compare our method with state-of-the-art methods recently published on this dataset. 

The quantitative results for ShanghaiTech-A are listed in Table~\ref{table:1}. We collect results of the state-of-the-art approaches from their original published papers.
It can be seen that our PDANet has achieved an MAE of \(58.5  \) and an MSE of \(93.4 \) in the experiment. 
Our proposed method also exhibits significant advantages over many top ranked methods such as PaDNet~\cite{tian2019padnet}, ADCrowdNet~\cite{liu2019adcrowdnet}, HA\_CNN~\cite{sindagi2019ha}, and SPN~\cite{chen2019spn}. On the ShanghaiTech-B dataset, our proposed PDANet has achieved an MAE of \(7.1 \) and an MSE of \(10.9 \), both are better than those of the state-of-the-art results. These results suggest that our proposed PDANet is able to cope with sparse and dense scenes, thanks to the combination of the pyramid module as mentioned in Sect.~\ref{sec:pfe} and the two-branch DAD as described in Sect.~\ref{sec:dad}. Because of these, our proposed model can distinguish the crowd level of the input scene and analyze the crowd accordingly for better estimation. 

\subsection{Experimental Results on the WorldExpo10 Dataset}
\label{sec:ewe}

The WorldExpo10 dataset~\cite{zhang2015cross} is another large-scale crowd counting benchmark dataset. 
During the Shanghai WorldExpo 2010, 1,132 video clips were captured by $108$  surveillance cameras to produce this large dataset. We follow the standard procedures~\cite{zhang2015cross} to do expriments on this dataset.
Table~\ref{table:1} also provides MAE results based on five different scenes. 
The best-performing state-of-the-art methods are CAN~\cite{liu2019context}, ADCrowdNet~\cite{liu2019adcrowdnet}, and  PACNN~\cite{shi2019revisiting} with an average MAE less than 8. 
However, as shown in the table, our proposed PDANet has achieved an average MAE of \(6.0\), which suppresses the-state-of-the-art results with a margin of $1.4$ over the results achieved by CAN~\cite{liu2019context}. 
Furthermore, our PDANet yields the lowest MAE of \(4\) out of all \(5\) scenes with an MAE values equal to 1.8, 9.1, 7.3, and 2.2, respectively. 
As it is demonstrated, the overall performance of our PDANet across various scenes is superior compared with the-state-of-the-art approaches.

\subsection{Experimental Results on the UCF Dataset}
\label{sec:eucf}

The UCF CC 50~\cite{idrees2013multi} is one of the most challenging data sets in crowd counting research area due to its limited number of training images and significant variation in the number of people within the datasets (from 94 to 4,543 across images). We choose the setting similar to the ShanghaiTech-A~\cite{zhang2016single} setting for generating ground truth density maps. Table~\ref{table:1} shows that our PDANet outperforms the state-of-the-art models by a significant margin. 
We achieve an MAE of $119.8$ with an MSE of $159$, which is about $35$ percent better than  PaDNet~\cite{tian2019padnet}, the best-performing benchmark model. 
In our experiments, we observe that our PDANet is able to estimate the number of people accurately in all subsets. Overall, it can be concluded that our proposed PDANet can work well on both sparse and dense scenarios. We also explore the results in detail at the Ablation Study In Section  \text{II} of the supplementary document.

\subsection{Experimental Results on the UCSD Dataset}
\label{sec:eucsd}

The UCSD dataset ~\cite{chan2008privacy} is another dataset that we conduct experiments on. In the experiments, we use Frames 601 through 1400 for training and the remaining out of 2000 for testing. 
Table~\ref{table:1} shows the MAE and MSE results obtained on this dataset. Compared with nine currently best approaches tested on this low crowed dataset, the proposed PDANet achieves the second best results with an MAE of \( .93 \) and an MSE of \( 1.21 \), which are very close and very comparative to the best results from the PaDNet model. We believe that the results is good enough taking into account that extreme resizing (image size in UCSD dataset is \(238 \times 158\)) is needed for PDANet as we mentioned in Sec.~\ref{sec:da} to work in our model. 

\section{Conclusion}

In this work, we introduced a novel deep architecture called Pyramid Density-Aware Attention-based network (PDANet) for crowd counting. 
The PDANet has incorporated pyramid features and attention modules with a density-aware decoder to address the huge density variation within the crowded scenes. 
The proposed PDANet has utilized a classification module for passing the pyramid features to the most suitable decoder branch to provide more accurate crowd counting with two-scale density maps. 
To aggregate these density maps, we took the benefit of the sigmoid function and produced a gating mask for producing the final density map. 
Extensive experiments on various benchmark datasets demonstrated the performance of our PDANet in terms of robustness, accuracy, and generalization. Our approach is able to achieve better performance on almost all of the major crowd counting datasets over the state-of-the-art methods, especially in UCF 50 with more than $35$ percent immediate improvements in the results.


\bibliographystyle{IEEEtran}

\bibliography {refs}

\ifCLASSOPTIONcaptionsoff
  \newpage
\fi

\end{document}